\documentclass[11pt,a4paper]{article}
\usepackage[hyperref]{emnlp2020}
\usepackage{times}
\usepackage{latexsym}

\usepackage{todonotes}
\usepackage{booktabs}
\usepackage{multirow}
\usepackage{amsmath}
\usepackage{amssymb}
\usepackage{colortbl}
\usepackage{xspace}
\usepackage{pifont}
\usepackage{graphicx}
\usepackage{textcomp}
\usepackage{microtype}
\usepackage{xspace}
\usepackage{bbm}

\aclfinalcopy 
\def\aclpaperid{2650} 

\newcommand{\pegc}{PEGASUS$_\textrm{CNN/DM}$\xspace}
\newcommand{\pegx}{PEGASUS$_\textrm{XSUM}$\xspace}
\newcommand{\bartc}{BART$_\textrm{CNN/DM}$\xspace}
\newcommand{\bartx}{BART$_\textrm{XSUM}$\xspace}

\title{Understanding Neural Abstractive Summarization Models \\ via Uncertainty}

\author{Jiacheng Xu \quad\quad Shrey Desai \quad\quad Greg Durrett \\
  Department of Computer Science \\
  The University of Texas at Austin\\
  \texttt{ \{jcxu,gdurrett\}@cs.utexas.edu \quad shreydesai@utexas.edu }
  }

\date{}

\begin{document}
\maketitle

\begin{abstract}
An advantage of seq2seq abstractive summarization models is that they generate text in a free-form manner, but this flexibility makes it difficult to interpret model behavior. 
In this work, we analyze summarization decoders in both blackbox and whitebox ways by studying on the entropy, or uncertainty, of the model's token-level predictions. For two strong pre-trained models, PEGASUS \cite{pegasus} and BART \cite{lewis-2019-bart} on two summarization datasets, we find a strong correlation between low prediction entropy and where the model copies tokens rather than generating novel text. The decoder's uncertainty also connects to factors like sentence position and syntactic distance between adjacent pairs of tokens, giving a sense of what factors make a context particularly selective for the model's next output token. Finally, we study the relationship of decoder uncertainty and attention behavior to understand how attention gives rise to these observed effects in the model. We show that uncertainty is a useful perspective for analyzing summarization and text generation models more broadly.\footnote{Code is available at \url{https://github.com/jiacheng-xu/text-sum-uncertainty}}

\end{abstract}

\section{Introduction}

Recent progress in abstractive summarization has been fueled by the advent of large-scale Transformers pre-trained on autoregressive language modeling objectives \cite{hoang-etal-2019,khandelwal-etal-2019,lewis-2019-bart,pegasus}. Despite their strong performance on automatic metrics like ROUGE \cite{lin-2004-rouge}, abstractive models are not as straightforward and interpretable as their extractive counterparts. Free-form generation in these models also leads to serious downstream errors, such as factual inconsistencies with the input document \cite{CaoEtAl2018fact,kryscinski-etal-2019-factuality,wang-etal-2020-qags,durmus-etal-2020-feqa,goyal-durrett-2020-factuality}. Although the interpretability of NLU models has been extensively studied \cite{ribeiro2016should,ghaeini-etal-2018-interpreting,jain-wallace-2019-attention,desai2020calibration}, summarization models specifically have not received similar attention, with analysis efforts often focused on datasets and evaluation \cite{kryscinski-etal-2019-neural}.


In this work, we focus on interpreting and understanding abstractive summarization models through the lens of decoder uncertainty, or the entropy of decisions during generation. While uncertainty in generation has been studied from the perspective of data \cite{ott2018analyzing}, sampling \citep{fan-etal-2018-hierarchical,holtzman2019curious}, and training  \cite{correia2019adaptively,kang2020improved}, it is underutilized as a technique for analysis and inspection of generation systems. We study two prominent summarization models, PEGASUS \cite{pegasus} and BART \cite{lewis-2019-bart}, fine-tuned on two English summarization datasets, CNN/Daily Mail \cite{hermann-2015-cnndm} and XSum \cite{narayan-2018-xsum}, to understand model behavior in each setting.

First, by comparing $n$-grams between the input document and generated summaries, we establish two coarse types for decoded tokens, \textit{copy} and \textit{generate} \cite{see-2017-ptrgen}. We find that the entropy of the generation decision correlates with whether the model is copying or generating, as well as where in the sentence the token is. This paints a picture of certain contexts being more restrictive from the standpoint of generation, particularly early in sentences where a model has not ``decided'' what to copy yet, and illustrates the interaction of content selection and lexical choice. 
Second, we extend this analysis by looking at how uncertainty relates to the syntax of the generated sentence: whether uncertainty connects to syntactic notions of surprisal \cite{roark-etal-2009-deriving} and how the entropy varies across certain syntactic productions.
Finally, we derive a way to quantify decoder attention by aggregating distinct self-attention heads, revealing the correlation between the attention entropy and prediction entropy, and investigating the correspondence between the prediction entropy and the fraction of the past and future decoded tokens.

Taking this analysis together, we find that the abstractiveness of reference summaries fundamentally changes model behavior: the extractive nature of CNN/DM makes most of its decisions low entropy and copy-oriented while the model maintains higher uncertainty on XSum, yielding more abstractive summaries. More broadly, we show that uncertainty is a simple but effective tool to characterize decoder behavior in text generation. 

\section{Model and Experimental Setup}
Our experiments use PEGASUS \cite{pegasus} and BART \cite{lewis-2019-bart}, two state-of-the-art seq2seq pre-trained models. We use the \textit{large} version of these two models, which have 16 and 12 Transformer layers, respectively. Both models have pre-training objectives tailored somewhat to this problem domain: seq2seq modeling for denoising (BART) or infilling of masked-out sentences (PEGASUS). We directly use the pre-trained models from \citet{Wolf2019HuggingFacesTS}.\footnote{Specifically, \texttt{google/pegasus-cnn$\_$dailymail}, \texttt{google/pegasus-xsm}, \texttt{facebook/bart-large-cnn}, and \texttt{facebook/bart-large-xsum} for PEGASUS and BART on these two datasets.}


As reported in the original papers and measured by ROUGE-1/2/L \cite{lin-2004-rouge}, PEGASUS achieves 44.17/21.47/41.11 on CNN/DM \cite{hermann-2015-cnndm} and 47.21/24.56/39.25 on XSum \cite{narayan-2018-xsum}, and BART achieves 44.16/21.28/40.90 and 45.14/22.27/37.25.

\paragraph{Entropy.} Entropy is a standard measure of uncertainty in a probabilistic distribution. Given a discrete random variable $X$ with all possible outcomes $x_1, \cdots, x_n$, the entropy of $X$ is defined as $H(X) = - \sum_{i=1}^{n}P(x_i)\log P(x_i)$.

For pre-trained Transformers, the domain of the predictions (the vocabulary) is large and also differs between models. The vocabulary sizes for PEGASUS and BART are 96,103 and 50,265,\footnote{Note that entropy generally increases as the variable's domain grows: a uniform distribution over 10,000 outcomes has entropy 9.21, while a uniform distribution over 100,000 outcomes has entropy 11.51.} and the prediction distribution is usually long-tailed. To combat this, nucleus sampling \citep{holtzman2019curious} is used to sample from only the top $1-p$ most probable outcomes (the nucleus) to avoid generating very unlikely tokens. To more fairly compare models with different vocabulary sizes, and to better reflect the actual sampling distribution, we therefore compute all entropy values in this work over the nucleus distribution. That is, we sort the prediction distribution $P(x_i)$ in descending order and get a minimal set of tokens where $V^{\text{min}} = \{ x | \sum_{x_i \in V^{\text{min}}} P(x_i) \geq p \}$. Then we re-normalize the distribution as follows:
\begin{equation}
  P'(x_i) =
    \begin{cases}
      \frac{P(x_i)}{p'} & \text{if $x_i \in V^{\text{min}}$}\\
      0 & \text{otherwise}.
    \end{cases}       
\end{equation}
where the cumulative probability $p' = \sum_{x_i \in V^{\text{min}}} P(x_i)$. We use $p=0.95$ for all experiments. The entropy $H(X)$ is computed based on the new distribution $P'(x_i)$.


\section{Model Uncertainty during Generation}
\label{sec:model-uncertainty}

In this section, we analyze and compare the prediction uncertainty from different models and different datasets by inspecting entropy values during generation, allowing us to localize uncertainty to certain positions in a decoded sentence. A principle factor that past work has investigated is the amount of copying in abstractive summarization models \cite{see-2017-ptrgen,paulus-2018-reinforced}.
We first aim to understand how decisions to copy document content or generate new text are reflected in the model's uncertainty.

One complicating factor is that while BART and PEGASUS both exhibit a mix of copying and novel generation, they do not have an explicit copy operation like in past models and so these behaviors are more difficult to define. We first separate generation decisions by bigrams that appear in the input document (existing bigrams) or whether they are free-form generations (novel bigrams).\footnote{Bigrams are defined based on tokens rather than wordpieces, and so may consist of more than two generation steps.}

\begin{figure}[t!]
\centering
\small
\includegraphics[width=0.48\textwidth]{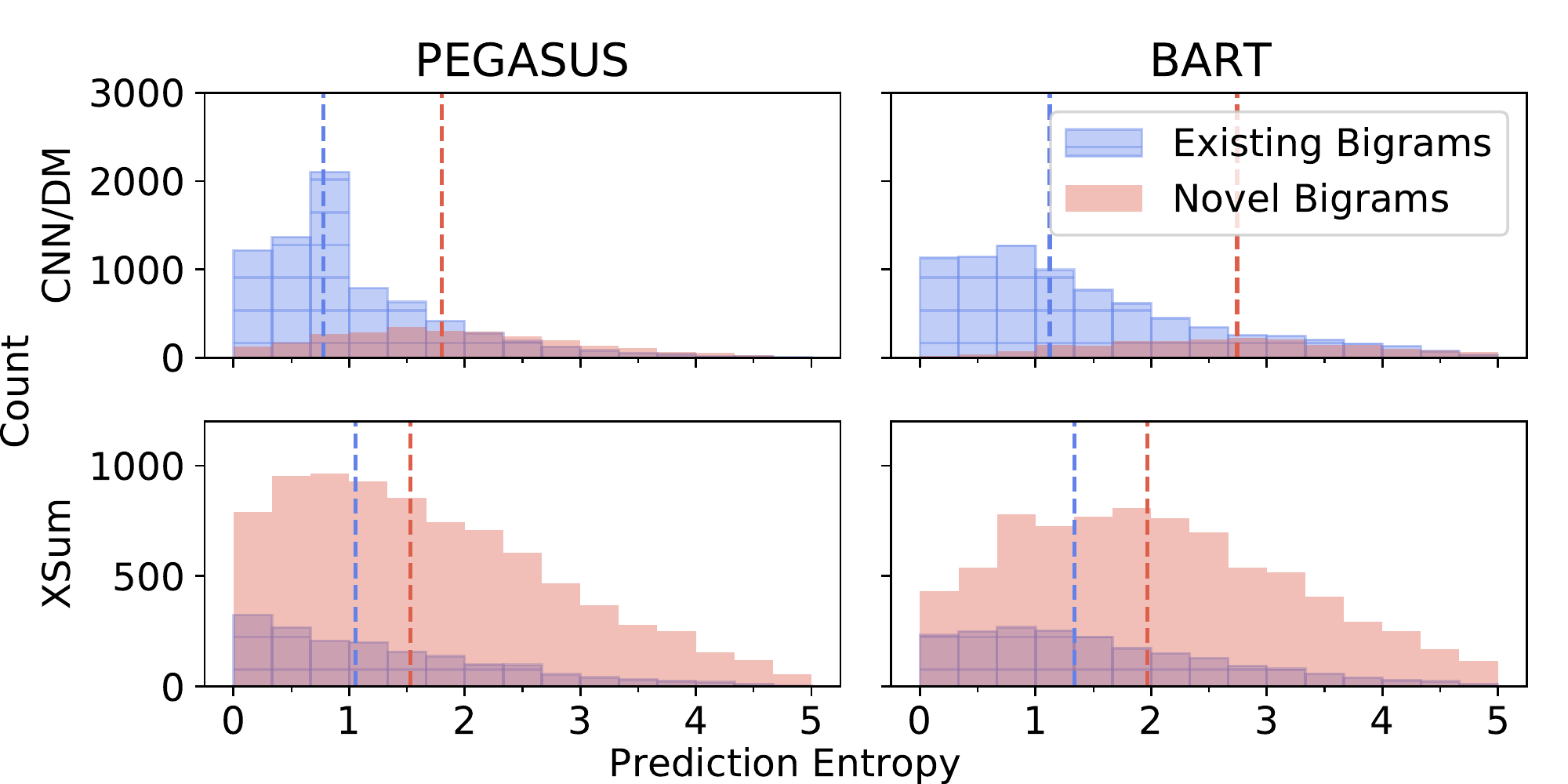}
\caption{Next token entropies computed on 10K generation steps from \pegc, \pegx, \bartc and \bartx respectively, broken into two cases: an \emph{Existing Bigram} means the bigram just generated occurs in the input document, while a \emph{Novel Bigram} is an organic model generation. These cases are associated with low entropy and high entropy actions, respectively. The x-axis shows the entropy (truncated at 5), and the y-axis shows the count of bigram falling in each bin. The dashed lines indicate the median of each distribution. 
}
\label{fig:ent_bigram}
\end{figure}

Figure~\ref{fig:ent_bigram} shows a histogram of model entropies broken down by these two categories. Most notably, \textbf{there is a strong correlation between copy-like behavior and the entropy of the model's prediction distribution}. On CNN/DM, we see that low entropy decisions are largely those generating existing bigrams, and conversely, existing bigrams are usually generated with low entropy. New bigrams are generated with a broad range of high entropy values, and are much more frequent on XSum.
These results align with our manual analysis of these summaries: \pegc and \bartc summaries largely consist of spans from the input document with minor compression while \pegx and \bartx summaries involve stitching together disparate concepts and paraphrasing key details. This reflects a corresponding divergence in the gold summaries, where CNN/DM summaries are far more extractive than those in XSum.

Critically, though the entropy distributions are dissimilar across the two datasets, we see regularities among the approximate \textit{copy} and \textit{generate} operations: on CNN/DM and XSum, the median entropy values of using existing bigrams are 0.95 and 1.20, respectively, and for generating new bigrams, 2.27 and 1.75. 

With this connection between entropy and copying behavior, we make the following additional observations based on Figures \ref{fig:ent_bigram} and \ref{fig:relpos_ent}:

\begin{figure}[t!]
\centering
\small
\includegraphics[width=0.48\textwidth]{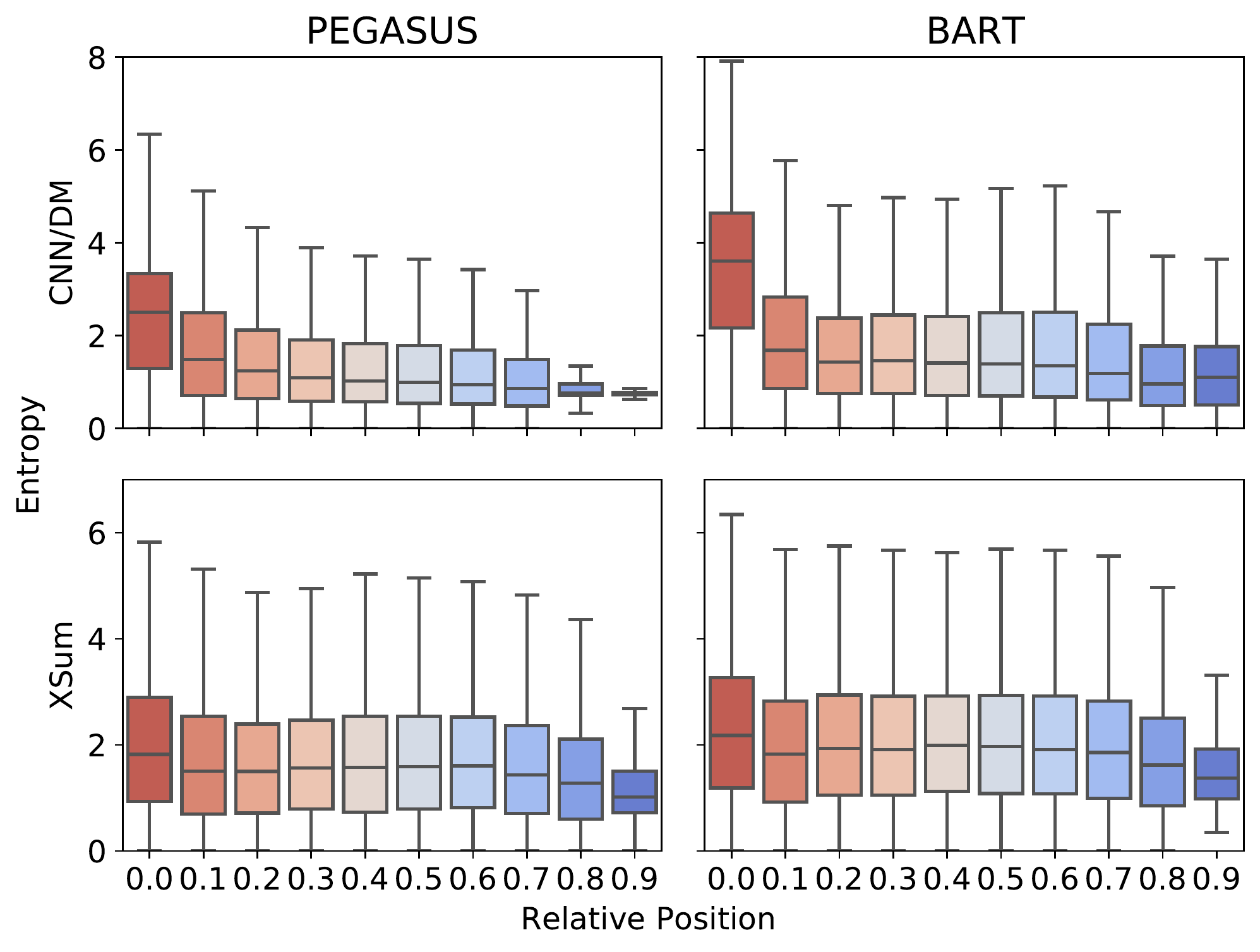}
\caption{Prediction entropy values by relative sentence positions. For example, 0.0 indicates the first 10\% of tokens in a sentence, and 0.9 is the last 10\% of tokens.
\pegc and \bartc make highly uncertain decisions to start, but then entropy decreases, suggesting that these models may be copying based on a sentence prefix. Entropies on XSum are more constant across the sentence.}
\label{fig:relpos_ent}
\end{figure}

\paragraph{Entropy varies across token positions, especially on CNN/DM.} In Figure~\ref{fig:relpos_ent}, we depict a different view of entropy, looking at the decoding process as it progresses through each sentence. Across both CNN/DM and XSum, models are most uncertain at the beginning of the sentence and least uncertain at the end of the sentence. However, the rate at which entropy drops off is quite different: on CNN/DM, the entropy after decoding 20\% of tokens falls below 2, while the entropies on XSum only begin to considerably drop after decoding 80\% of tokens. Our manual analysis suggests the following characterization: \textbf{to generate each sentence on CNN/DM, the model makes some high-entropy decisions to identify a sentence and begin to copy its prefix, followed by a series of low entropy decisions to copy that sentence's content.} On XSum, which is highly abstractive and features single sentence summaries, content planning and generation are less clearly decoupled. 

\paragraph{PEGASUS copies and generates more tokens with entropy $< 1$.} BART and PEGASUS report similar ROUGE results on CNN/DM, but these models do not place the same distributions over summaries. PEGASUS has more low-entropy copying decisions, and its start-of-sentence entropies are also significantly lower (Figure~\ref{fig:relpos_ent}). This suggests that it is more confident than BART in selecting content to discuss next. There are also more low-entropy generation decisions, particularly on XSum.

\section{Entropies of Syntactic Productions}

Having observed connections between sentence position and entropy, we now flesh out this analysis from the lens of syntax, focusing in particular on uncertainty at constituent boundaries. From our PEGASUS generations on CNN/DM and XSum, we obtain constituency parses for each summary sentence using the Berkeley Neural Parser \cite{kitaev-klein-2018-constituency} and explore connections between syntax and uncertainty in more depth.

\begin{figure}[t!]
\centering
\small
\includegraphics[width=0.48215\textwidth]{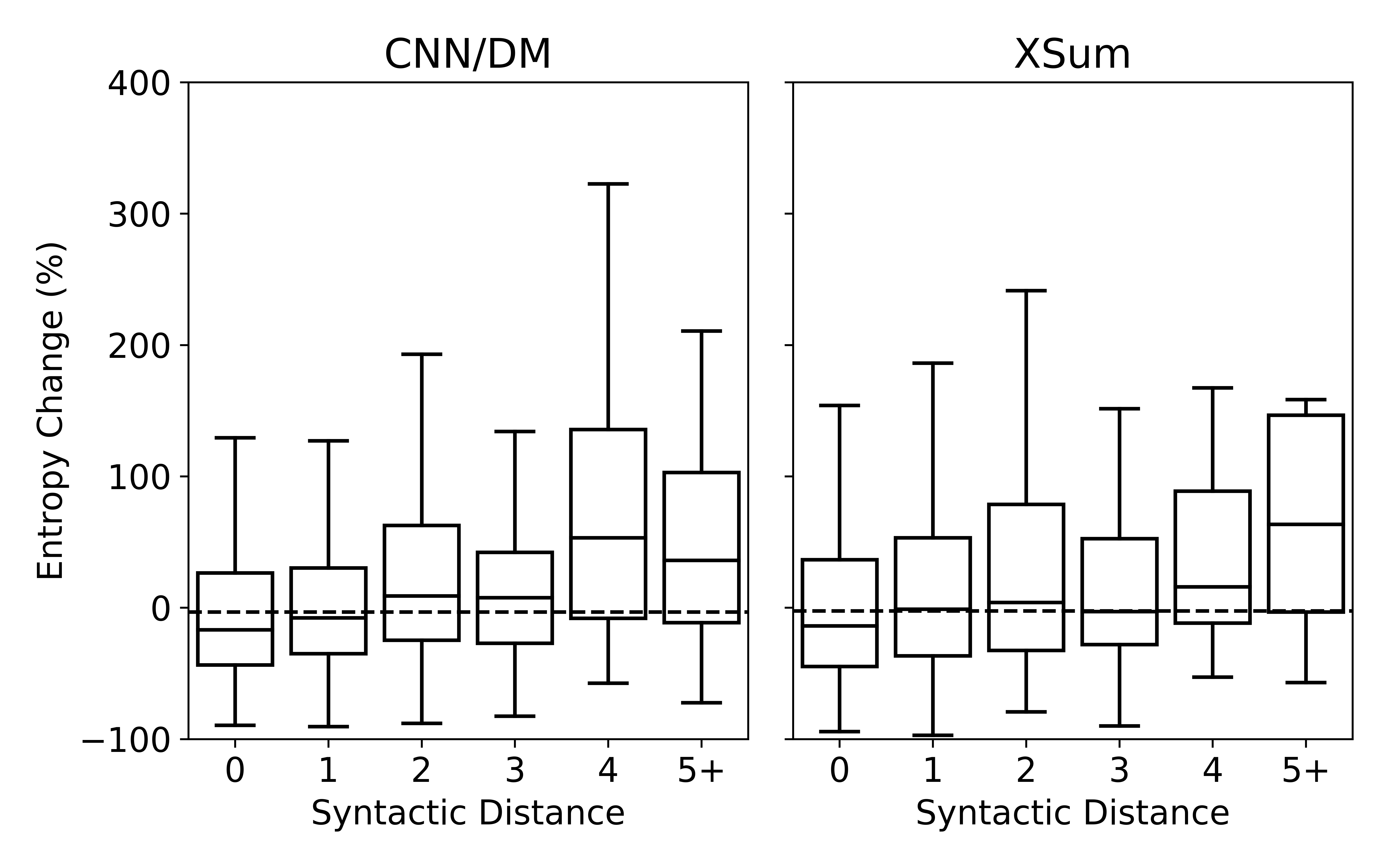}
\caption{Correlating syntactic distance between neighboring tokens with the entropy change in those tokens' generation decisions for PEGASUS summaries. The median entropy change is depicted as a dashed black line. At points of high syntactic distance, the model's behavior is less restricted by the context, correlating with higher entropy.}
\label{fig:pegasus-syntactic-distance}
\end{figure}

\paragraph{Low and high entropy decisions can be localized to constituent span boundaries.} Parsing has long been used to explain psycholinguistic notions of surprisal \cite[inter alia]{hale-2001-probabilistic,roark-etal-2009-deriving}, which are in turn related to uncertainty under a language model. In our case, uncertainty about \emph{generating} a text is a different notion than uncertainty when a reader is \emph{processing} it. Hence, rather than looking at an incremental parser's behavior, we instead look at a simpler notion of syntactic distance \cite{shen-2018-neural}, or the number of left and right parentheses between $w_t$ and $w_{t+1}$ in a linearized constituency tree. Our hypothesis is that when these words exhibit high syntactic distance, this word boundary is a ``choice point'' where the model may be less restricted in what it can choose to generate next.

Figure~\ref{fig:pegasus-syntactic-distance} shows the correlation between syntactic distance and the percent change in entropy between the adjacent tokens. On both CNN/DM and XSum, we see two patterns emerge: generating a token within the same immediate parent constituent (i.e., zero syntactic distance) is typically a \textit{certain} decision, while generating a token belonging to a new constituent is an increasingly \textit{uncertain} decision. From these results, we can draw a parallel to the copy vs. generate behavior established in Section~\ref{sec:model-uncertainty}; for example, generating \emph{York} after \emph{New} might be straightforward, perhaps due to a direct copy from the document, but generating a prepositional phrase might be more challenging due to the large search space of possible constructions or the higher chance that the model might delete this constituent.

\begin{table}[t]
\setlength{\tabcolsep}{4pt}
\small
\centering
\begin{tabular}{lp{4.2cm}}
\toprule
Production Rule & Example \\
\midrule
\texttt{NP $\rightarrow$ NP : NP} & [Arsenal vs Reading]$_{1.2}$ [:]$_{0.6}$ [the game that changed the game]$_{3.1}$ \\
\texttt{NP $\rightarrow$ NP , SBAR ,} & [driver]$_{0.5}$ [,]$_{0.4}$ [who has not been identified]$_{2.2}$ [,]$_{0.1}$ \\
\midrule
\texttt{NP $\rightarrow$ CD NN NNS} & [16]$_{0.07}$ [felony]$_{0.05}$ [counts]$_{0.01}$ \\
\texttt{NP $\rightarrow$ NNP CD} & [April]$_{0.04}$ [3]$_{0.1}$ \\
\bottomrule
\end{tabular}
\caption{Examples of specific NP productions with high entropy (top) and low entropy (bottom). The notation [Y]$_{H(Y)}$ implies the constituent $Y$ is generated with entropy $H(Y)$.}
\label{tab:production-examples}
\end{table}

\paragraph{Low entropy spans are often short, specific units of information.} We also investigate the average entropy of spans within a rule production to uncover what types of spans are likely to elicit certainty or uncertainty during generation. In Table~\ref{tab:production-examples}, we see qualitatively that productions with low average entropy productions are short extracts of document content, such as \textit{16 felony counts}. These are largely factual, often containing cardinal values, and more likely to be copied. Within these constituents, the model is very certain about what to generate next, supporting the connection with low syntactic distance.


\section{Understanding Decoder Self-Attention}


While we have analyzed the model's predictions, we have not yet determined \emph{how} the different behaviors we see emerge from the context. 
Our goal is to explore what the encoder attention places its emphasis during generation and how it correlates with the prediction entropy.\footnote{In PEGASUS and BART models, the encoder and decoder attention during decoding are two separate distributions where the encoder attention looks at the encoding context and the decoder attention attends to the previously decoded tokens. In this paper we chiefly examine the encoder attention to understand how the model references the input document.}

\begin{figure}[t!]
\centering
\small
\includegraphics[width=0.48215\textwidth]{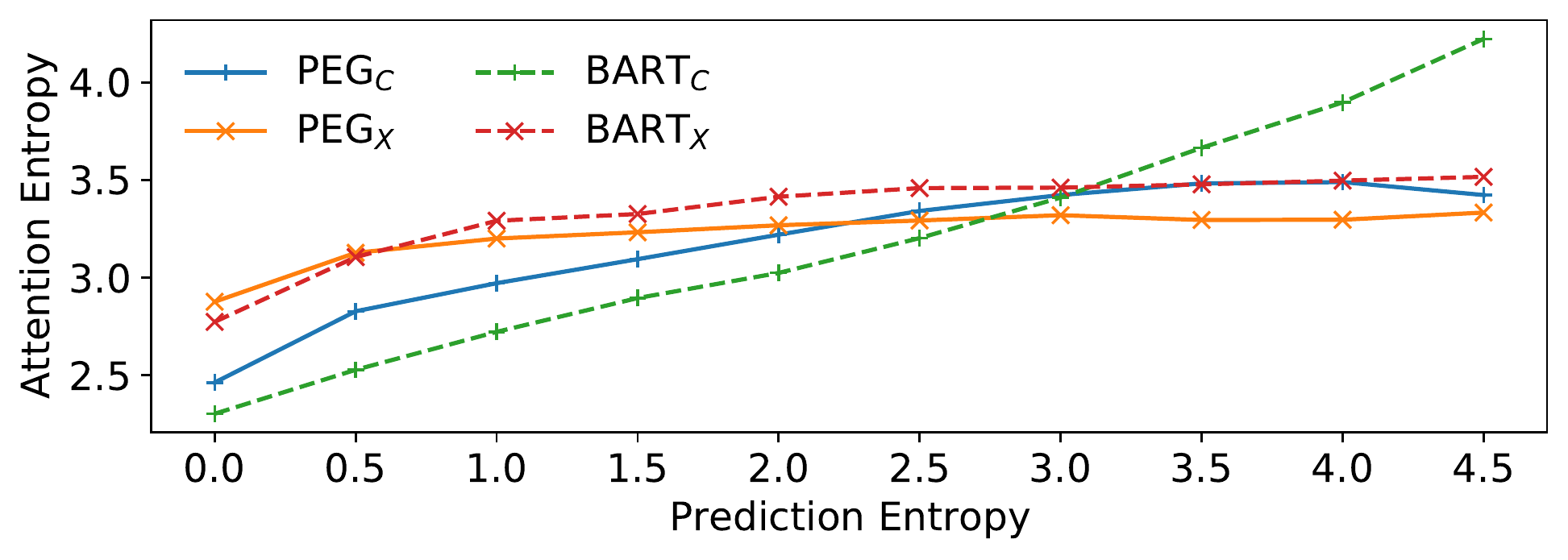}
\caption{Correlation between attention entropy and prediction entropy of PEG(ASUS) and BART on C(NN/DM) and X(Sum).
We compute the mean value of the attention entropy within each bucket of prediction entropy. 
The uncertainty of attention strongly correlates with the entropy of the model's prediction. }
\label{fig:attn-ent}
\end{figure}

\paragraph{Blocking Low-information Tokens.} Analyzing the inner workings of attention in Transformers is challenging \cite{clark-2019-bert,kovaleva-2019-revealing}, particularly because many heads are useless, redundant, or noisy, and they frequently attend to low-information tokens such as end-of-sentence markers or periods.
Inspired by tf-idf \cite{joachims1997probabilistic}, we propose a method to compute a set of tokens most meaningfully attended to by the model.
If a token in the encoding document is attended to across many time steps (like a word appearing in many documents in tf-idf), we want to disregard it in our analysis.


Let $T$ denote the number of decoder timesteps and $L$ be the length of the source document.
We compute an aggregate attention matrix $S \in \mathbb{R}^{T \times L}$ by summing the attentions across all heads and all layers. We then compute a count of how often each token is attended to above a threshold $q$: $f_l = \sum_{t=1}^{T} [\mathbbm{1}(s_{tl} \geq q)]$ and discard the attention values on tokens with the highest $f$ score. In practice we discard 5\% of tokens from the source document. 


\paragraph{Attention Entropy.}

One natural question we can ask is whether there is a connection between entropy of the attention distribution and entropy of the decoder's prediction. 
This relationship is shown in Figure~\ref{fig:attn-ent}, where each point represents the mean attention entropy within the corresponding prediction entropy bucket. The attention entropy is especially low where the prediction entropy ranges from 0 to 0.5.
For cases with prediction entropy greater than 1.5, the attention entropy saturates and no longer grows with the prediction entropy except the \bartc. While attention entropy is probably not ``causing'' the low decoder entropy per se, nevertheless decoder entropy provides a lens into the inner workings of the Transformer model.

\begin{figure}[t!]
\centering
\small
\includegraphics[width=0.48215\textwidth]{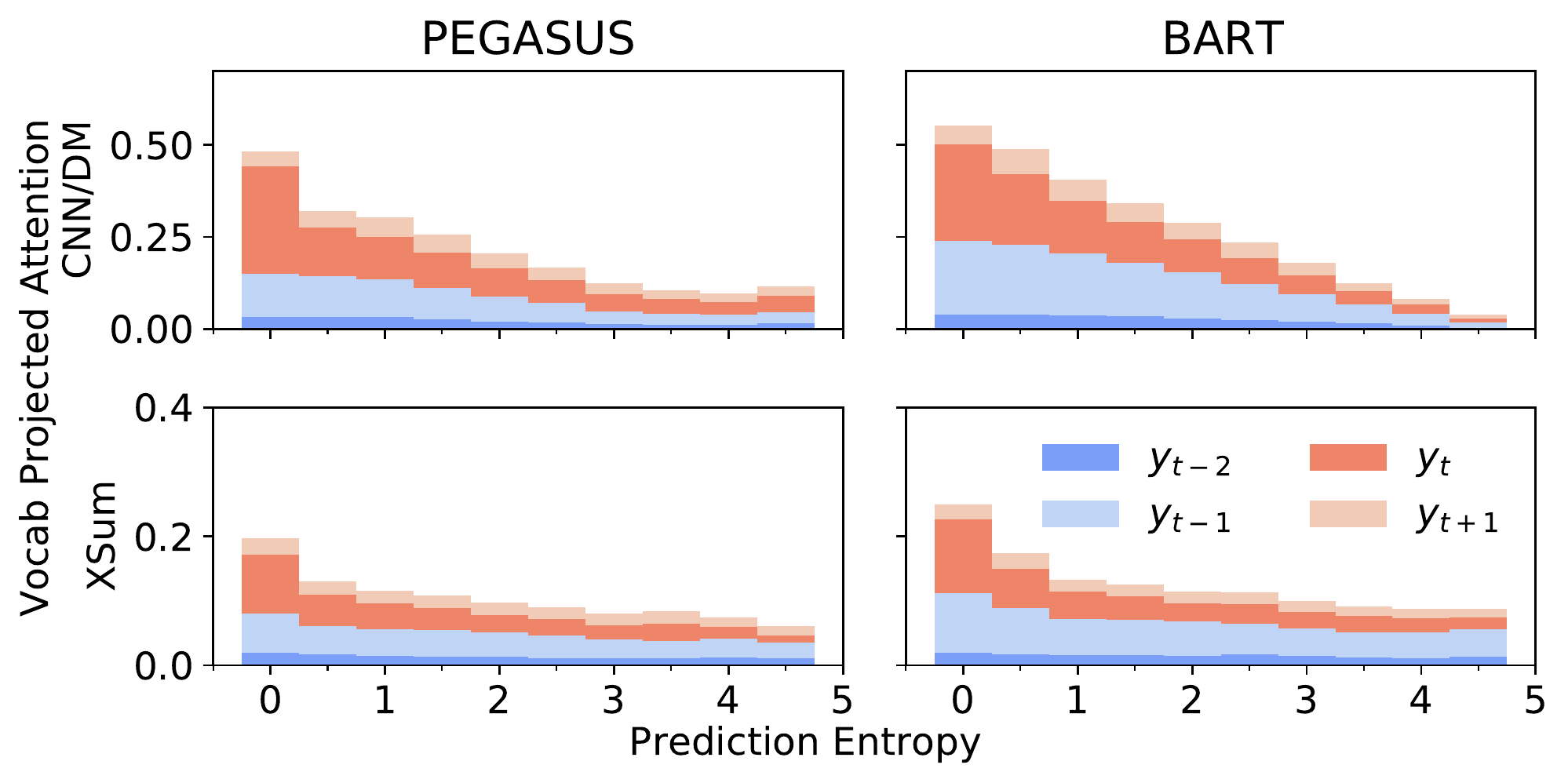}
\caption{Vocabulary projected attention attending to the last input $y_{t-2}$, current input $y_{t-1}$, current output $y_t$, and next output $y_{t+1}$. When the prediction entropy is low, the attention mostly focus a few tokens including the current input $y_{t-1}$ and current output $y_t$.}
\label{fig:attn}
\end{figure}

\paragraph{Projecting Attention to Vocabulary.}
We hypothesize that low decoder entropies may arise if the model is heavily attending to certain relevant tokens, particularly the (about to be predicted) token $y_t$ of time step $t$ and the input token of this time step $x_t$, equivalent to $y_{t-1}$. 
For the predicted token $y_t$, we compute the vocabulary projected attention value $\sum_{l=1}^{L} \mathbbm{1}[\text{token}_l = y_t] s_{tl}  $ where we accumulate the attention of  all of the occurrences of the specified token $y_t$ in the document. The higher the value, the more attention put to the encoding token(s) which are predicted for this time step during decoding.
We can define the value for last time step input $y_{t-2}$, current time step input $y_{t-1}$, and the not-yet-decoded token $y_{t+1}$ for next time step. 

We show the relationship between the vocabulary projected attention and the prediction entropy in Figure~\ref{fig:attn}.
Visualizations for both models and both datasets show that when the prediction entropy is low, the attention focuses heavily on a few tokens including the current input token and the current token to predict. This suggests a potential mechanism where the model indexes into the source document by attending to $y_{t-1}$, then strongly identifies and ``reads off'' $y_t$ as the next token to generate.

\section{Conclusion}

This work analyzes pre-trained summarization models via uncertainty, or the entropy of decoding decisions. We pursue several lines of inquiry: uncertainty can help us understand copying document spans vs.~generating novel text, the behavior of models in different syntactic environments, and coarse properties of the model's attention distribution. All of these give insight into what conditions most heavily restrict the model's generation: generating an observed bigram (copying), low syntactic distance, and attention which can easily identify decoder context in the source document. We believe this approach can power future analyses of pre-trained text generation systems.

\section*{Acknowledgments}

This work was partially supported by NSF Grant IIS-1814522, NSF Grant SHF-1762299, a gift from Salesforce Inc, and an equipment grant from NVIDIA. The authors acknowledge the Texas Advanced Computing Center (TACC) at The University of Texas at Austin for providing HPC resources used to conduct this research. Results presented in this paper were obtained using the Chameleon testbed supported by the National Science Foundation.
Thanks as well to the anonymous reviewers for their helpful comments.

\bibliography{emnlp2020}
\bibliographystyle{acl_natbib}








\end{document}